\documentclass[10pt,twocolumn,letterpaper]{article}

\usepackage{iccv}
\usepackage{times}
\usepackage{epsfig}
\usepackage{graphicx}
\usepackage{amsmath}
\usepackage{amssymb}

\usepackage{tablefootnote}
\usepackage{subcaption}
\usepackage{graphicx}
\usepackage{verbatim}
\usepackage{color}
\usepackage{booktabs}
\graphicspath{{images/}}
\usepackage{float}
\usepackage{enumitem}

\usepackage[pagebackref=true,breaklinks=true,letterpaper=true,colorlinks,bookmarks=false]{hyperref}

\usepackage{array}
\newcolumntype{P}[1]{>{\centering\arraybackslash}p{#1}}
\newcolumntype{M}[1]{>{\centering\arraybackslash}m{#1}}

\newcommand{\inputsurface}{x}
\newcommand{\surface}{\hat{x}}

\newcommand{\geo}{g}
\newcommand{\semantic}{s}

\newcommand{\Gsurface}{\mathcal{G}_{\hat{x}}}
\newcommand{\Ggeo}{\mathcal{G}_{g}}
\newcommand{\Gsemantic}{\mathcal{G}_{s}}

\newcommand{\Dsurface}{\mathcal{D}_{x}}
\newcommand{\Dsemantic}{\mathcal{D}_{s}}

\newcommand{\GTgeo}{g_\text{gt}}
\newcommand{\GTsemantic}{s_\text{gt}}

\newcommand{\plist}[1]{(\textit{#1})}

\newcommand{\figref}[1]{Fig.~\ref{#1}}
\newcommand{\secref}[1]{Sec.~\ref{#1}}
\newcommand{\tabref}[1]{Table~\ref{#1}}

\newcommand{\dimthree}[3]{{#1}$\times${#2}$\times${#3}}

\newcommand{\RemoveAboveCaptionTab}{1pt}
\newcommand{\RemoveAboveCaption}{-1pt}
\newcommand{\RemoveBelowCaption}{-3pt}
\newcommand{\RemoveAboveParagraph}{-10pt}

\usepackage[breaklinks=true,bookmarks=false]{hyperref}

\iccvfinalcopy 

\def\iccvPaperID{5704} 

\ificcvfinal\fi

\begin{document}

\title{ForkNet: Multi-branch Volumetric Semantic Completion\\from a Single Depth Image}

\author{
Yida Wang\textsuperscript{1}, 
David Joseph Tan\textsuperscript{2}, 
Nassir Navab\textsuperscript{1}, 
Federico Tombari\textsuperscript{1,2}\\
\textsuperscript{1}Technische Universit\"at M\"unchen ~~~~~
\textsuperscript{2}Google Inc. 
}

\makeatletter
\g@addto@macro\@maketitle{
  \begin{figure}[H]
  \setlength{\linewidth}{\textwidth}
  \setlength{\hsize}{\textwidth}
  \centering
\includegraphics[width=1.0\linewidth]{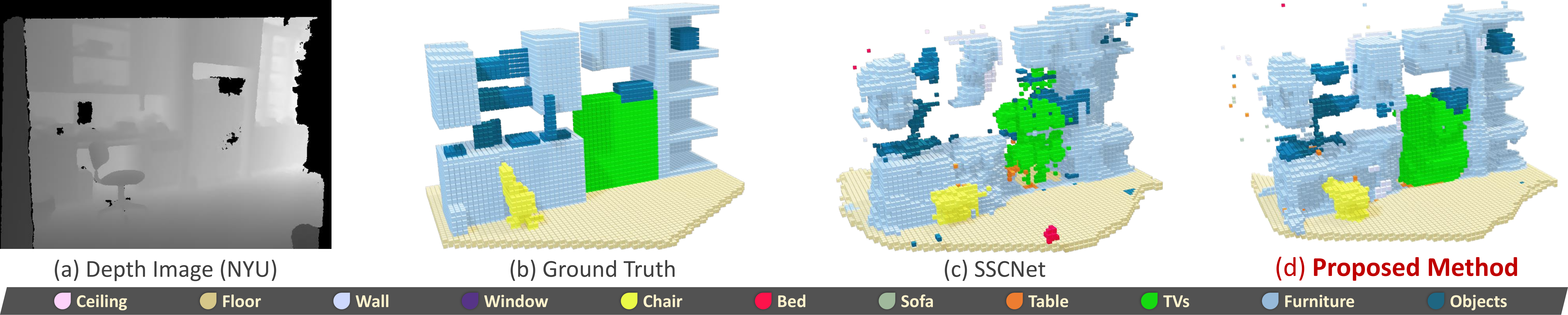}
	\setlength{\abovecaptionskip}{-8pt}
	\setlength{\belowcaptionskip}{3pt}
	\caption{Our 3D semantic completion model (right-most) generates realistic yet accurate volumetric scene representations from a single depth image (left-most) affected by occlusion and noise, even if acquired from a real depth sensor.
	}
	\label{fig:teaser}
	\end{figure}
}
\makeatother

\maketitle
\ificcvfinal\thispagestyle{empty}\fi

\begin{abstract}
	We propose a novel model for 3D semantic completion from a single depth image, based on a single encoder and three separate generators used to reconstruct different geometric and semantic representations of the original and completed scene, all sharing the same latent space. 
	To transfer information between the geometric and semantic branches of the network, we introduce paths between them concatenating features at corresponding network layers. 
	Motivated by the limited amount of training samples from real scenes, an interesting attribute of our architecture is the capacity to supplement the existing dataset by generating a new training dataset with high quality, realistic scenes that even includes occlusion and real noise. 
	We build the new dataset by sampling the features directly from latent space which generates a pair of partial volumetric surface and completed volumetric semantic surface. 
	Moreover, we utilize multiple discriminators to increase the accuracy and realism of the reconstructions. We demonstrate the benefits of our approach on standard benchmarks for the two most common completion tasks: semantic 3D scene completion and 3D object completion.
\end{abstract}

\section{Introduction}

The increasing abundance of depth data, thanks to the widespread presence of depth sensors on devices such as robots and smartphones, has recently fostered big advancements in 3D processing for augmented reality, robotics and scene understanding, unfolding new applications and technology that relies on the geometric rather than just the appearance information.  
Since 3D devices sense the environment from one specific viewpoint, the geometry that can be captured in one shot is only partial  due to occlusion caused by foreground objects as well as self-occlusion from the same object. 

As for many applications, this partial 3D information is insufficient to robustly carry-out 3D tasks such as object detection and tracking or scene understanding. A recent research direction has emerged that leverages deep learning to ``complete" the depth images acquired by a 3D sensor, \ie filling in the missing geometry that the sensor could not capture due to occlusion. The capability of deep learning to determine a latent space that captures the global context from the training samples proved useful in regressing completed 3D scenes and 3D shapes even when big portion of the geometry are missing~\cite{dai2017shape, dai2018scancomplete,rethage2018deep, song2017semantic,  yang20173d}. Also, some of these approaches have been extended to jointly learn how to infer geometry and semantic information, in what is referred to as semantic 3D scene completion~\cite{dai2018scancomplete,song2017semantic, wang3dv}. 
Nevertheless, current approaches are still limited by different factors, including the difficulty of regressing fine and sharp details of the completed geometry, as well as to generalize to shapes that significantly differ from those seen during training. 

\begin{figure}[!t]
	\centering
	\includegraphics[width=1.0\linewidth]{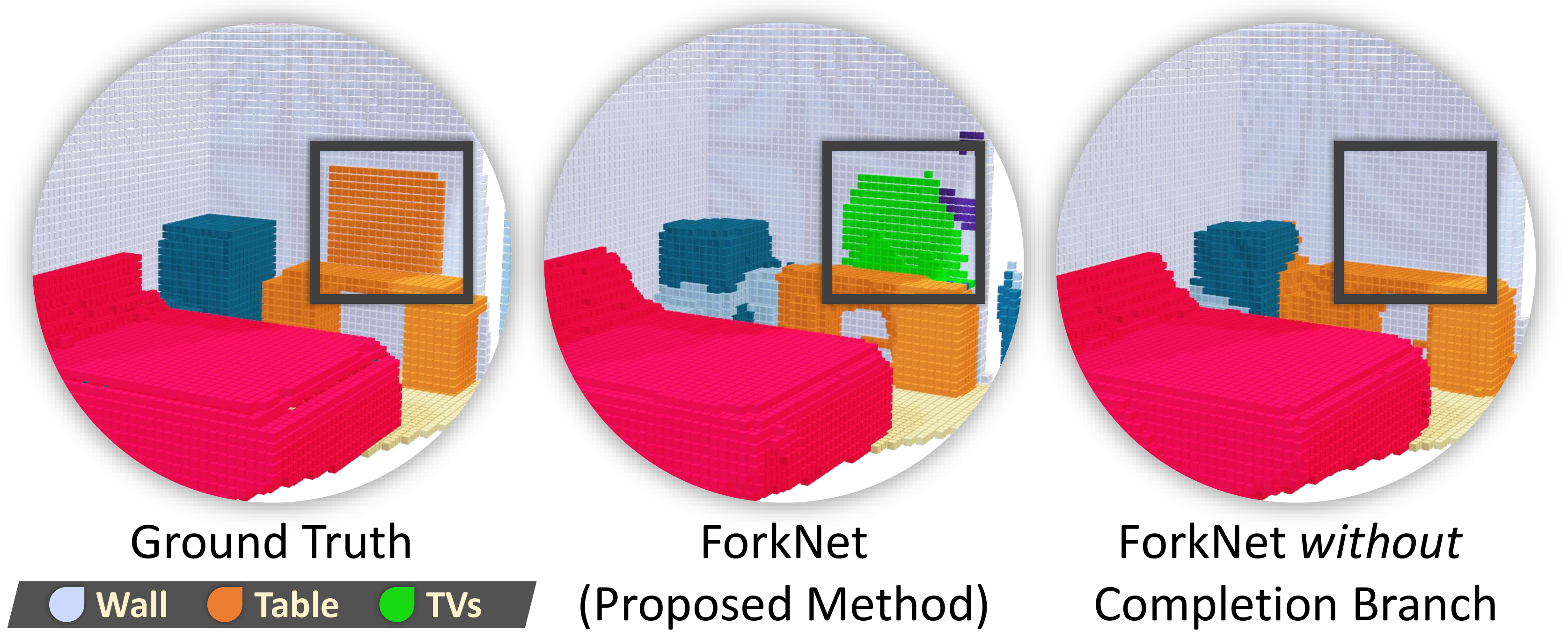}
	\setlength{\abovecaptionskip}{\RemoveAboveCaption}
	\setlength{\belowcaptionskip}{\RemoveBelowCaption}
	\caption{This figure shows the ground truth reconstruction where we notice the incorrect labels from SUNCG~\cite{song2017semantic} dataset on the TVs, \ie enclosed in the black box.
	}
	\label{fig:annotation}
\end{figure}

%

In this work, we aim to tackle 3D completion from a single depth image based on a novel learned model that relies on a single encoder and multiple generators, each trained to regress a different 3D representation of the input data: \plist{i}~a voxelized depth map, \plist{ii}~a geometric completed volume, \plist{iii}~a semantic completed volume. This particular architecture aims at two goals. The first is to supplement the lack of paired input-output data, \ie a depth map and the associated completed volumetric scene, with novel pairs directly generated from the latent space, \ie by means of \plist{i} and \plist{iii}. The second goal is to overcome a common limitation of available benchmarks that provide imprecise semantic labels, by letting the geometric completion remain unaffected from it, \ie by means of \plist{i} and \plist{ii}. 
By means of specific connections between corresponding neural layers in the different branches, we let the semantic completion model be conditioned on geometric reconstruction information, this being beneficial to generate accurate reconstructions with aligned semantic information.

Overall, the proposed learning model uses a mix of supervised and unsupervised training stages which leverage the power of generative models in addition to the annotations provided by benchmark datasets. Additionally, we propose to further improve the effectiveness of our generative model by employing discriminators able to increase the accuracy and realism of the produced output,  
yielding completed scenes with high level details even in the presence of strong occlusion, as witnessed by \figref{fig:teaser} that reports an example from a real dataset (NYU~\cite{SilbermanECCV12}). 

Our contributions can be summarized as follows: \plist{i} a novel architecture, dubbed ForkNet, based on a single encoder and three generators built upon the same shared latent space, useful to generate additional paired training samples; 
\plist{ii} the use of specific connections between generators to let geometric information condition and drive the completion process over the often imprecise ground truth annotations (see \figref{fig:annotation}); and, \plist{iii} the use of multiple discriminators to regress fine details and realistic completions. 
We demonstrate the benefits of our approach on standard benchmarks for the two most common completion tasks: semantic 3D scene completion and 3D object completion. For the former, we rely on SUNCG~\cite{song2017semantic} (synthetic) and NYU~\cite{SilbermanECCV12} (real). For the latter, instead, we test on ShapeNet~\cite{chang2015shapenet} and 3D-RecGAN~\cite{yang2018dense}. Notably, we outperform the state of the art for both scene reconstruction and object completion on the real dataset.

\begin{figure*}[!t]
	\centering
	\includegraphics[width=1.0\linewidth]{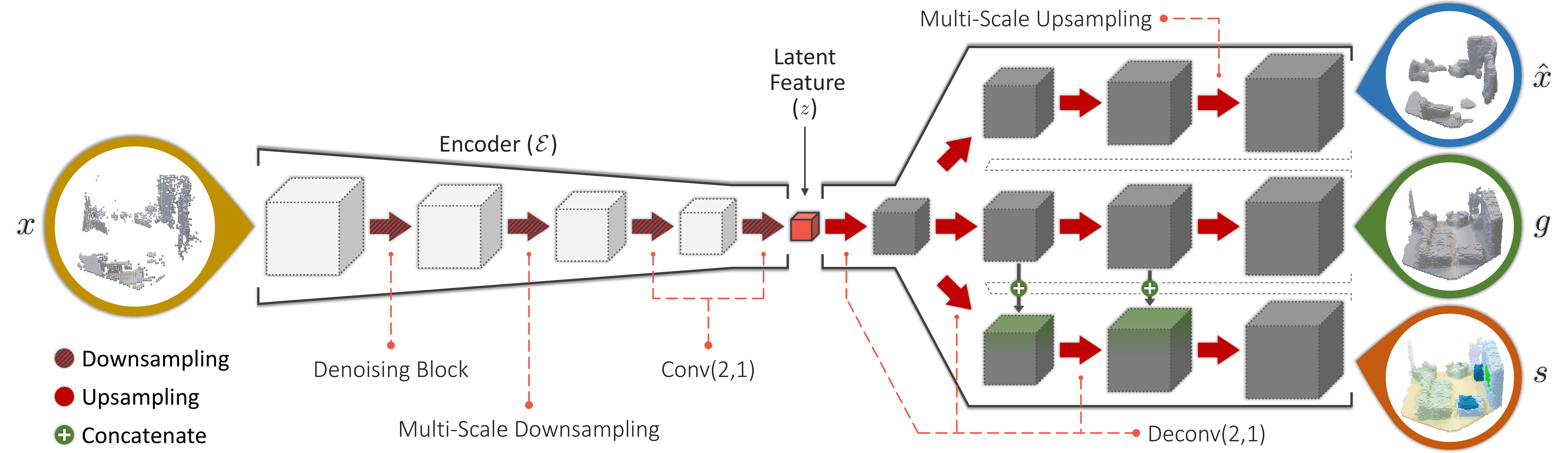}
	\setlength{\abovecaptionskip}{\RemoveAboveCaption}
	\setlength{\belowcaptionskip}{\RemoveBelowCaption}
	\caption{ForkNet -- the proposed volumetric network architecture for semantic completion relies on a shared latent space encoded from SDF volume $\inputsurface$ reconstructed from the input depth image. The two decoding paths are trained to generate, respectively, incomplete surface geometry ($\surface$), completed geometric volume ($\geo$) and completed semantic volumes ($\semantic$). 
	}
	\label{fig:architecture}
\end{figure*}

\section{Related work}
\paragraph{Semantic scene completion.}
3D semantic scene completion starts from a depth image or a point cloud to provide an occlusion-free 3D reconstruction of the visible scene within the viewpoint's frustrum while labeling each 3D element with a semantic class from a pre-defined category set.
Scene completion could be in principle achieved by exploiting simple geometric cues such as plane consistency~\cite{monszpart2015rapter} or object symmetry~\cite{kim2012acquiring}. Moreover, meshing approaches such as Poisson reconstruction~\cite{kazhdan2013screened} as well as purely geometric works~\cite{firman2016structured} can also be employed for this goal. 

Recent approaches suggested to leverage deep learning to predict how to fill-in occluded parts in a globally coherent way with respect to the training set. 
%
SSCNet~\cite{song2017semantic} carries out semantic scene completion from a single depth image using dilated convolution~\cite{yu2015multi} to capture 3D spatial information at multiple scales. They rely on a volumetric representation to represent both input and output data. 
Based on SSCNet, VVNet~\cite{guo2018view} applies view-based 3D convolutions as a replacement for SDF back-projections, this resulting more effective in extracting geometric information from the input depth image. 
SaTNet~\cite{liu2018nips} relies on the RGB-D images. They initially predict the 2D semantic segments with the RGB. The depth image then back-projects the semantically labelled pixels to a 3D volume which goes through another architecture for 3D scene completion.
ScanComplete~\cite{dai2018scancomplete} also targets semantic scene completion but, instead of starting from a single depth image, they assume to process a large-scale reconstruction of a scene acquired via a consumer depth camera. They suggest a coarse-to-fine scheme based on an auto-regressive architecture~\cite{reed2017parallel}, where each level predicts the completion and the per-voxel semantic labeling at a different voxel resolution. 
The work in~\cite{wang3dv} proposes to use GANs for the task of semantic scene completion from a single depth image. In particular, it proposes to use adversarial losses applied on both the output and latent space to enforce realistic interpolation of scene parts. 
The work in~\cite{wang3dv} proposes to use GANs for the task of semantic scene completion from a single depth image. In particular, it proposes to use adversarial losses applied on both the output and latent space to enforce realistic interpolation of scene parts. 
Partially related to this field, the work in \cite{tulsiani2018factoring} leverages input object proposals in the form of 2D bounding boxes to extract the layout of a 3D scene from a single RGB image, while estimating the pose of the objects therein. A similar task is tackled by \cite{geiger2015joint} starting from an RGB-D image.

\vspace{\RemoveAboveParagraph}
\paragraph{Object completion.}
3D object completion aims at obtaining a full 3D object representation from either a single depth or RGB image. While several RGB-based approaches have been recently proposed \cite{choy20163d, fan2017point, wu2017marrnet}, in this section, we will focus only on those based on depth images as input since they are more related to the scope of this work. 
The work in \cite{rethage2018deep} uses a hybrid architecture based on a CNN and an autoencoder to learn completing 3D shapes from a single depth map. 
3D-RecGAN~\cite{yang2018dense,yang20173d} proposes to complete an observed object from a single depth image using a network based on skip connections~\cite{ronneberger2015u} between the encoder and the generator so to fetch more spatial information from the input depth image to the generator.
3D-EPN~\cite{dai2017shape} performs shape completion based on a latent feature concatenated with object classification information via one-hot coding, so that this additional semantic information could drive an accurate extrapolation of the missing shape parts.
Han \etal~\cite{han2017high} complete shapes with multiple depth images fused via LSTM Fusion~\cite{li2016lstm} and process the fused data using a 3D fully convolutional approach.
MarrNet~\cite{wu2017marrnet} reconstructs the 3D shape by applying reprojection consistency between 2.5D sketch and 3D shape.

\begin{figure*}[!ht]
  \begin{minipage}[c]{0.75\textwidth}
    \includegraphics[width=\textwidth]{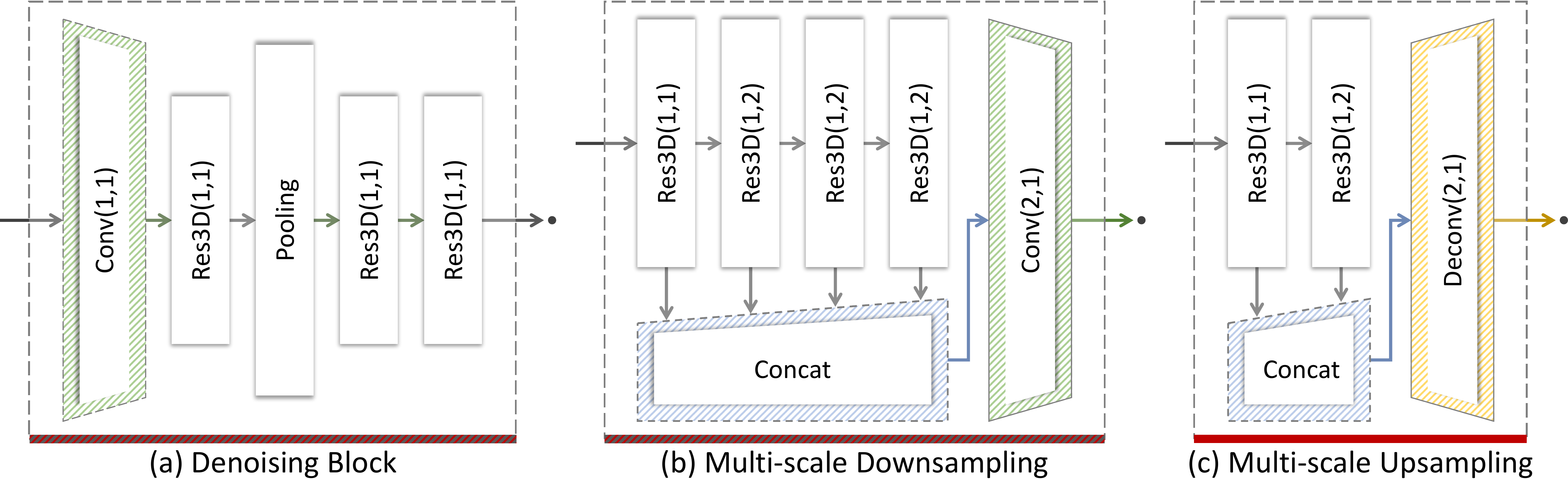}
  \end{minipage}\hfill
  \begin{minipage}[c]{0.23\textwidth}
	\caption{(a-b) Downsampling and (c) upsampling convolutional layers in our architecture (see \figref{fig:architecture}).
	Note that the two parameters $(s,d)$ in all the functions are the stride and dilation while the kernel size is set to 3.
\label{fig:arch_micro}}
  \end{minipage}
\end{figure*}

\vspace{\RemoveAboveParagraph}
\paragraph{GANs for 3D shapes.}
Although the use of GANs for 3D semantic scene completion tasks is almost an unexplored territory, GANs have been frequently employed in recent proposals for the task of learning a latent space for 3D shapes, useful for object completion as well as for tasks such as object retrieval and object part segmentation.  
For instance, 
3D-VAE-GAN~\cite{wu2016learning} trains a volumetric GAN in an unsupervised way from a dataset of 3D models, so to be able to generate realistic 3D shapes by sampling the learned latent space.
ShapeHD~\cite{shapehd} tackles the difficult problem of reconstructing 3D shapes from a single RGB image and suggests to overcome the 2D-3D ambiguity by adversarially learning a regularizer for shapes.
PrGAN~\cite{gadelha20173d} learns to generate 3D volumes in an unsupervised way, trained by a discriminator that distinguishes whether 2D images projected from a generated 3D volume are realistic or fake.
3D-ED-GAN~\cite{8237514} transforms a coarse 3D shape into a more complete one using a Long Short-term Memory (LSTM) Network by interpreting 3D volumes as sequences of 2D images.

\section{Proposed semantic completion}



Taking the depth image as input, we reconstruct the visible surface by back-projecting each pixel onto a voxel of the volumetric data. 
Denoted as $\inputsurface$, we represent the surface reconstruction from the depth image as a signed distance function (SDF)~\cite{osher2003signed} with $n_l \times n_w \times n_h$ voxels such that the value of the voxel approaches zero when it is closer to the visible surface.
 
Our task then is to produce the completed reconstruction of the scene with a semantic label for each voxel. 
Having $N$ object categories, the class labels are assigned as $\mathcal{C}=\{c_i\}_{i=0}^N$ where $c_0$ is the empty space. Thus, denoted as $\semantic$, we represent the resulting semantic volume as a one-hot encoding~\cite{luc2016semantic} with $N+1$ dimensional feature.
Similarly, we define $\geo$ as the completed reconstruction of the scene without the semantic information by setting $N$ to 1.

\subsection{Model architecture}

We assemble an encoder-generator architecture~\cite{wu2016learning} that builds the completed semantic volume from the partial scene derived from a single depth image. 
As illustrated in \figref{fig:architecture}, the encoder $\mathcal{E}(\cdot)$ is composed of 3D convolutional operators where the spatial resolutions are decreased by a factor of two in each layer.
In effect, this continuously reduces the volume into its simplest form, denoted by the latent feature $z$ such that $z=\mathcal{E}(\inputsurface)$.

In detail, the encoder is composed of four downsampling operators. 
The first aims at denoising~\cite{song2017semantic} the SDF volumes as illustrated in \figref{fig:arch_micro}(a).
This involves a combination of a 3D convolutional operator, several 3D ResNet blocks~\cite{hara2018can}, denoted as $\text{Res3D}(s,d)$ where $s$ is the stride while $d$ is the dilation, and a pooling layer. 
The second layer aims at including different objects in the scene even with varying sizes by concatenating the output of four sequentially connected 3D ResNet blocks in \figref{fig:arch_micro}(b). 
Consequently, the information from the smaller objects are captured on the first $\text{Res3D}(\cdot,\cdot)$ while the larger object are captured on the subsequent blocks. 
Notably, the first block is parameterized with a dilation of 1 while the other three with dilations of 2.
The concatenated result is then downsampled by a 3D convolutional operator.
In the final two layers, we further downsample the volume with 3D convolutional operators 
until we form the latent feature with a size of $16\times5\times3\times5$.

Branching from the same latent feature, we design three generators that reconstructs:
%
\begin{enumerate}[label=(\roman*),leftmargin=*,noitemsep]
	\item the \emph{SDF volume} ($\surface$) which, with respect to $\inputsurface$, formulates as an autoencoder; 
	\item the \emph{completed volume} ($\geo$) which focuses on reconstructing the geometric structure of the scene;
	and,
    \item the \emph{completed semantic volume} ($\semantic$) which is the desired outcome. 
\end{enumerate}
We assign these generators as the functions $\Gsurface(\cdot)$, $\Ggeo(\cdot)$ and $\Gsemantic(\cdot)$, respectively. 
Notably, we distinguish $\inputsurface$, which is the SDF volume obtained from the input depth image, from $\surface$, which is the inferred SDF volume obtained from the generator.
The structure of each generator is composed of 3D deconvolutional operators that increases the spatial resolution by two in each layer. 

While the first 3 convolutional upsampling layers in the generators are composed of 3D deconvolutional operators as shown in \figref{fig:architecture}, the last layer is a multi-scale upsampling which is sketched in \figref{fig:arch_micro}(c). 
This layer is similar to the multi-scale downsampling of the encoder where the goal is to consider the variation of sizes from different objects.
In this case, we concatenate the results of two sequentially connected 3D ResNet blocks then end with a 3D deconvolution operator. 
With the same operations as the other generators, the generator that builds the completed semantical volume $\Gsemantic$ additionally incorporates the data from the generator of the geometric scene reconstruction $\Ggeo$ as shown in \figref{fig:architecture} by concatenating the results from the second and the third layers.
Since the resulting $\surface$, $\geo$ and $\semantic$ have different number of channels, only the dimension of the output from the deconvolutional operator in the last layer changes for each structure.

\begin{figure*}[!ht]
	\centering
	\includegraphics[width=1.0\linewidth]{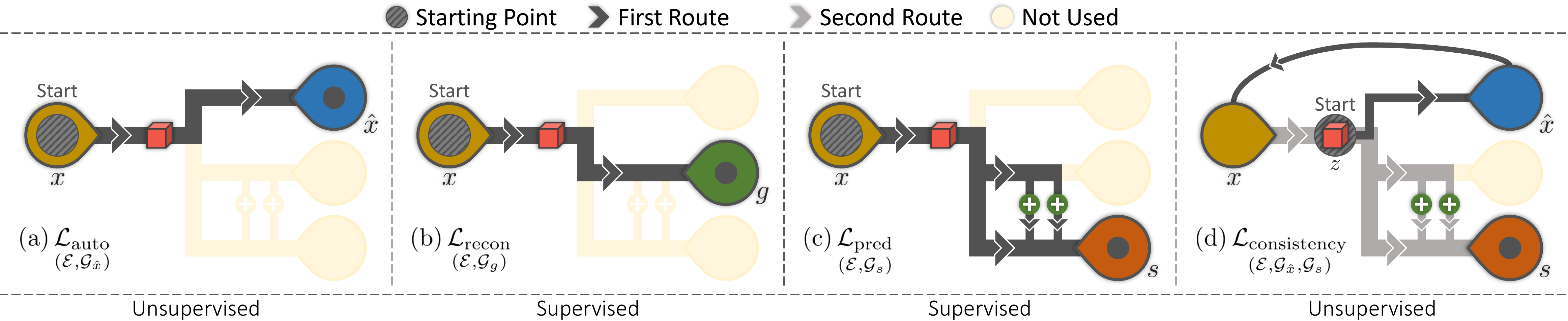}
	\setlength{\abovecaptionskip}{\RemoveAboveCaption}
	\setlength{\belowcaptionskip}{\RemoveBelowCaption}
	\caption{Graphical models of the 4 data flows (and the associated loss terms) used during training and derived from \figref{fig:architecture}.
	}
	\label{fig:path}
\end{figure*}

Giving a holistic perspective, we can simplify the sketch of the architecture in \figref{fig:architecture} to \figref{fig:path} by plotting the relation of the variables $\inputsurface$, $\surface$, $\geo$, $\semantic$ and $z$.
When we focus on certain structures, we notice that we have 
an autoencoder that builds an SDF volume in \figref{fig:path}(a), 
the reconstruction of the scene in \figref{fig:path}(b) and 
the volumetric semantic completion in \figref{fig:path}(c), where all of these structures branch out from the same latent feature.
Later in \secref{sec:loss}, these plots are used to explain the loss terms in training.


The rationale of having multiple generators is twofold. First, in contrast to the typical encoder-decoder architecture, we introduce the connection that relates the two generators. Taking the output from the $\Gsurface$ in each layer, we concatenate the results to the data from $\Gsemantic$ as shown in \figref{fig:architecture}. By establishing this relation, we incorporate the SDF reconstruction from the $\Gsurface$ into the semantic completion in order to capture the geometric information of the observed scene.

Second, the latent feature can generate a pair of SDF and completed semantic volumes. Through this set of paired volumes, we can supplement the learning dataset in an unsupervised manner.
This becomes a significant component in evaluating the NYU dataset~\cite{SilbermanECCV12} in \secref{sec:eval_scene} where the amount of learning dataset is limited because, since they use a consumer depth camera to capture real scenes, annotation becomes difficult. 
However, evaluating on this dataset is more essential compared to the synthetic dataset because it brings us a step closer to real applications. 
Relying on this idea in \secref{sec:loss}, we propose an unsupervised loss term that optimizes the entire architecture based on its own learning dataset.

\vspace{\RemoveAboveParagraph}
\paragraph{Discriminators.}

Inspired by GANs~\cite{goodfellow2014generative, radford2015unsupervised}, we introduce the discriminator $\Dsurface$ that evaluates whether the generated SDF volumes from $\Gsurface$ are realistic or not by comparing them to the learning dataset.
Here, $\Dsurface$ is constructed by a sequentially connected 3D convolutional operators with the kernel size of \dimthree{3}{3}{3} and stride of 2. 
This implies that the resolution of the input volume is sequentially decreased by a factor of two after each operation. 
To capture the local information of the volume \cite{demir2018patch}, the results from $\Dsurface$ is set to a resolution of \dimthree{5}{3}{5}.

With a similar architecture as $\Dsurface$, we also introduce a second discriminator $\Dsemantic$ that evaluates the authenticity of the generated volume $\semantic$. Notably, the two discriminators are evaluated in the loss terms in \secref{sec:loss} to optimize the generators.

\subsection{Loss terms}
\label{sec:loss}

Leveraging on the forward passes of smaller architectures in \figref{fig:path}, we can optimize the entire architecture by simultaneously optimizing different paths.
We also optimize the architecture of the two discriminators that distinguishes whether the generated volumes are realistic or not. 
During training, the learning dataset is given by a set of the pairs $\{(\inputsurface, \GTsemantic)\}$, where we distinguish $\GTsemantic$ as the ground truth from the generated $\semantic$.
Note that the ground truth for the geometric completion $\GTgeo$ is the binarized summation of non-empty space in $\GTsemantic$ and an occupancy volume from the SDF surface.

\vspace{\RemoveAboveParagraph}
\paragraph{SDF autoencoder.}

Motivated to reconstruct as similar SDF volume from the generator $\Gsurface$ as the original input, we define the loss function 
\begin{align}
	\underset{(\mathcal{E},\Gsurface)}{\mathcal{L}_{\text{auto}}} 
	= \left\|\Gsurface(\mathcal{E}(\inputsurface)) - {\inputsurface}\right\|^2
	\label{eq:l_recon_sce}
\end{align}
for the autoencoder in \figref{fig:path}(a) in order to minimize the difference between the observed $\inputsurface$ and the inferred $\surface$.

\vspace{\RemoveAboveParagraph}
\paragraph{Geometric completion.}

In \figref{fig:path}(b), a conditional generative model combines the encoder $\mathcal{E}(\cdot)$ and the generator $\Ggeo(\cdot)$ in order to reconstruct the scene (\ie without the semantic labels). Since the reconstruction is a two-channel volume that represents the empty and non-empty category, we use a binary cross-entropy loss  
\begin{align}
	\underset{(\mathcal{E},\Ggeo)}{\mathcal{L}_{\text{recon}}} &= \sum_{i=0}^{1}(\epsilon(\Ggeo(\mathcal{E}(\inputsurface)), \GTgeo))
	\label{eq:l_recon}
\end{align}
to train the inference network,
where $\epsilon(\cdot,\cdot)$ is the per-category error 
\begin{align}
	\epsilon(q,r) = -\lambda r \log q - (1- \lambda)(1 - r)\log(1 - q)~.
	\label{eq:per_category_error}
\end{align}
In \eqref{eq:per_category_error}, $\lambda$, which ranges from 0 to 1, weighs the importance of reconstructing true positive regions in the volume. If $\lambda=1$, the penalty for the false positive predictions will not be considered; while, if $\lambda$ is set to 0, the false negatives will not be corrected.

\vspace{\RemoveAboveParagraph}
\paragraph{Semantic completion.}

Similar to \eqref{eq:l_recon},
in \figref{fig:path}(c), we train a conditional generative model that is composed of the encoder $\mathcal{E}(\cdot)$ and generator $\Gsemantic(\cdot)$ linking $\inputsurface$ and $\semantic$.
Hence, we also use a binary cross-entropy loss 
\begin{align}
	\underset{(\mathcal{E},\Gsemantic)}{\mathcal{L}_{\text{pred}}} &= \sum_{i=0}^{N}(\epsilon(\Gsemantic(\mathcal{E}(\inputsurface)), \GTsemantic))
	\label{eq:l_pred}
\end{align}
where $N$ is the number of categories in the semantic scene. 

\vspace{\RemoveAboveParagraph}
\paragraph{Discriminators on the architecture.}

In relation to the architecture, we use two discriminators to optimize the generators \cite{wu2016learning} through
\begin{align}
	\underset{\Gsurface}{\mathcal{L}_{\text{gen-}\surface}} &=  - \log{(\Dsurface(\Gsurface(z)))} \nonumber \\
	\underset{\Gsemantic}{\mathcal{L}_{\text{gen-}\semantic}} &= -\log{(\Dsemantic(\Gsemantic(z)))}~. \label{eq:l_generators}
\end{align}
%
%
In this manner, we optimize the two generative models including both the SDF encoder and the semantic scene generator by randomly sampling the latent features.
On the other hand, when we update the parameters of both discriminators, we optimize the loss functions 
\begin{align}
	\underset{(\Dsurface)}{\mathcal{L}_{\text{dis-}\inputsurface}} &= 
	-\log(\Dsurface(\inputsurface)) 
	-\log{(1 - \Dsurface(\Gsurface(z)))} \nonumber \\
	\underset{(\Dsemantic)}{\mathcal{L}_{\text{dis-}\semantic}} &= 
	-\log(\Dsemantic(\GTsemantic)) 
	-\log{(1 - \Dsemantic(\Gsemantic(z)))}~. \label{eq:l_discrim}
\end{align}
%
%
During training, we apply the set of equations in \eqref{eq:l_generators} and \eqref{eq:l_discrim} alternatingly to optimize the generators and the discriminators separately.
Note that we use the KL-divergence from the variational inference~\cite{ghahramani2000variational, hoffman2013stochastic} to penalize the deviation between the distribution of $\mathcal{E}(\inputsurface)$ and a normal distribution with zero mean and identity variance matrix. The advantage of such is the capacity to easily sample from the latent space in the generative model, which becomes helpful in the succeeding loss term.

\begin{figure}[!t]
	\centering
	\includegraphics[width=0.99\linewidth]{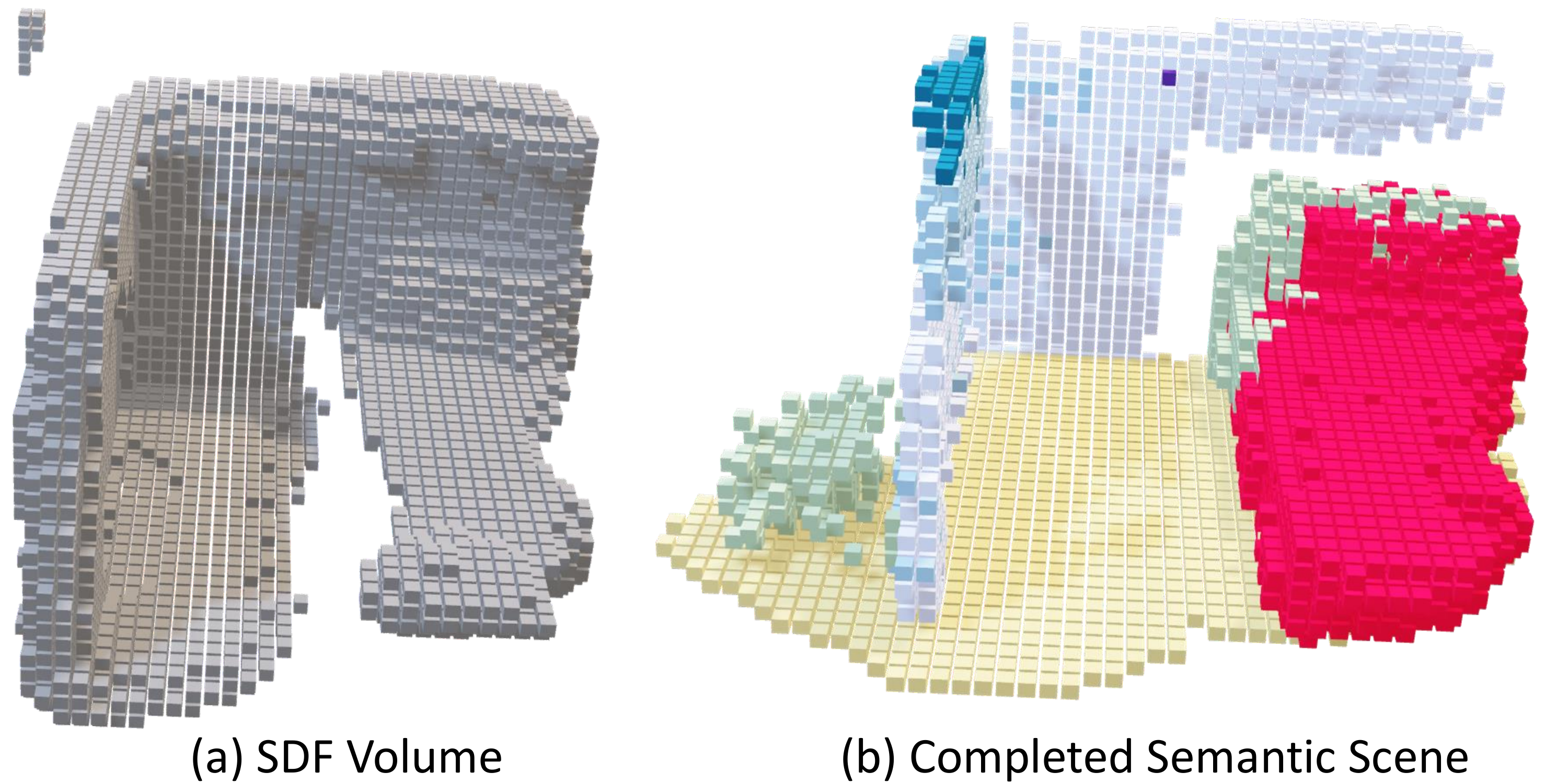}
	\setlength{\abovecaptionskip}{\RemoveAboveCaption}
	\setlength{\belowcaptionskip}{\RemoveBelowCaption}
	\caption{An example of the generated SDF volume and the corresponding completed semantic scene parameterized from the latent feature, which are used to supplement the existing learning dataset.}
	\label{fig:learn_data}
\end{figure}

\vspace{\RemoveAboveParagraph}
\paragraph{SDF-Semantic consistency.}

Since the generators are trained to produce SDF volumes and semantic scenes while being optimized to produce realistic data by the discriminator, we can build a new set of paired volumes to act as the learning dataset in order to supplement the existing one.
Thus, we propose to generate paired volumes directly from the latent feature in order to optimize the architecture in an unsupervised learning.

Exploiting the latent space, we reconstruct the set of pairs $\{(\Gsurface(z), \Gsemantic(z))\}$, where $z$ is randomly sampled from a Gaussian distribution centered on the average of latent features of a batch of samples. 
Following the inference model in \figref{fig:path}(c), we formulate a similar loss function as \eqref{eq:l_pred} but with the newly acquired data such that   
\begin{align}
	\underset{(\mathcal{E},\Gsurface,\Gsemantic)}{\mathcal{L}_{\text{consistency}}} &= \sum_{i=0}^{N}(\epsilon(\Gsemantic(\mathcal{E}(\Gsurface(z))), \Gsemantic(z)))~.
	\label{eq:l_consistency}
\end{align}
By drawing the data flow of the first term $\Gsemantic(\mathcal{E}(\Gsurface(z)))$ in \figref{fig:path}(d), we observe that the loss term in \eqref{eq:l_consistency} optimizes the entire architecture.

Interestingly, when we take a closer look at the newly generated pairs $\{(\Gsurface(z), \Gsemantic(z))\}$ in 
\figref{fig:learn_data},
we can easily notice the realistic results. The SDF volume in \figref{fig:learn_data}(a) considers missing regions due to the camera position while the semantic scene in \figref{fig:learn_data}(b) generates lifelike structures and reasonable positions of the objects in the scene (\eg the bed in red). 
By adding the newly generated pairs, we numerically show in \secref{sec:eval_scene} that there is a significant boost in performance when evaluating the NYU dataset~\cite{SilbermanECCV12} where the size of the learning dataset is small.

\paragraph{Optimization.} 

With all the loss terms given, achieving the optimum parameters in our architecture requires us to simultaneously minimize them.
We start by optimizing \eqref{eq:l_recon_sce}, \eqref{eq:l_recon}, \eqref{eq:l_pred} and \eqref{eq:l_generators} altogether. 
Then, the loss functions in \eqref{eq:l_discrim} for the two discriminators are optimized alternatively (\ie batch-by-batch) with \eqref{eq:l_recon_sce}, \eqref{eq:l_recon}, \eqref{eq:l_pred} and \eqref{eq:l_generators}.
In practice, we employ the Adam optimizer~\cite{kingma2014adam} with a learning rate of 0.0001.
For the data flows, \figref{fig:path}(a) and (d) are both unsupervised while \figref{fig:path}(b) and (c) are supervised. In addition, for the discriminators, \eqref{eq:l_generators} is unsupervised while \eqref{eq:l_discrim} is supervised.

\begin{table*}[!ht]
\centering
\resizebox{\textwidth}{!}
{\begin{tabular}{l|ccccccccccc|c}
	\toprule	
		\multicolumn{1}{c}{} 
	& ceil. & floor & wall & win. & chair & bed & sofa & table & tvs & furn. & objs. & \emph{Avg.} \\
	\midrule
		SSCNet~\cite{song2017semantic} (observed) & 97.7 & 94.5 & 66.4 & 30.0 & \textbf{36.9} & 60.2 & 62.5 & \textbf{56.3} & 12.1 & 46.7 & \textbf{33.0} & 54.2 \\
		\textbf{Proposed Method} (observed) & \textbf{98.2} & \textbf{96.9} & \textbf{67.8} & \textbf{37.4} & 35.9 & \textbf{72.9} & \textbf{69.6} & 48.8 & \textbf{20.5} & \textbf{48.4} & 32.4 & \textbf{57.2} \\
	\midrule
	Wang \etal~\cite{wang3dv} & 41.4 & 37.7 & 45.8 & 26.5 & 26.4 & 21.8 & 25.4 & 23.7 & 20.1 & 16.2 & 5.7 & 26.4 \\
	3D-RecGAN~\cite{yang2018dense} & 79.9 & 75.2 & 48.2 & 28.9 & 20.2 & 64.4 & 54.6 & 25.7 & 17.4 & 33.7 & 24.4 & 43.0 \\
	SSCNet~\cite{song2017semantic} & 96.3 & 84.9 & 56.8 & 28.2 & 21.3 & 56.0 & 52.7 & 33.7 & 10.9 & 44.3 & 25.4 & 46.4 \\
	VVNet~\cite{guo2018view} & \textbf{98.4} & \textbf{87.0} & 61.0 & {54.8} & \textbf{49.3} & \textbf{83.0} & 75.5 & \textbf{55.1} & {43.5} & \textbf{68.8} & \textbf{57.7} & \textbf{66.7} \\
	SaTNet~\cite{liu2018nips} & 97.9 & 82.5 & 57.7 & \textbf{58.5} & 45.1 & 78.4 & 72.3 & 47.3 & \textbf{45.7} & 67.1 & 55.2 & 64.3 \\
	\textbf{Proposed Method} & 95.0 & 85.9 & \textbf{73.2} & 54.5 & 46.0 & 81.3 & 74.2 & 42.8 & 31.9 & 63.1 & 49.3 & 63.4 \\ 
	-- \emph{without completion branch} & 94.1 & 83.5 & 68.2 & 49.6 & 43.1 & 80.5 & \textbf{77.7} & 41.8 & 33.8 & 61.7 & 51.7 & 62.3 \\
	-- \emph{without scene consistency} & 89.6 & 79.5 & 63.4 & 46.3 & 39.0 & 77.5 & 73.2 & 37.7 & 29.8 & 57.4 & 46.7 & 58.2 \\
	\bottomrule
	\end{tabular}
}
	\setlength{\abovecaptionskip}{\RemoveAboveCaptionTab}
\caption{Semantic scene completion results on the SUNCG test set with depth map for IoU (in \%). 
\label{tab:suncg_iou}
}
\end{table*}

\begin{table*}[!ht]
\centering
\resizebox{\textwidth}{!}
{\begin{tabular}{l|ccccccccccc|c}
	\toprule	
		\multicolumn{1}{c}{} 
	& ceil. & floor & wall & win. & chair & bed & sofa & table & tvs & furn. & objs. & \emph{Avg.} \\
	\midrule
		SSCNet~\cite{song2017semantic} (observed)& 37.7 & \textbf{91.9} & \textbf{75.4} & 64.0 & 29.0 & 51.1 & 63.3 & 43.7 & \textbf{29.7} & 73.3 & 54.5 & 50.8 \\
		\textbf{Proposed Method} (observed)& \textbf{41.5} & 90.8 & 69.6 & \textbf{54.8} & \textbf{27.7} & \textbf{53.1} & \textbf{66.3} & \textbf{44.4} & 27.1 & \textbf{74.7} & \textbf{57.5} & \textbf{55.2}  \\
	\midrule
		Lin \etal~\cite{lin2013holistic} (NYU only) & 0.0 & 11.7 & 13.3 & 14.1 & 9.4 & 29.0 & 24.0 & 6.0 & 7.0 & 16.2 & 1.1 & 12.0 \\
		3D-RecGAN~\cite{yang2018dense} & 35.3 & 70.3 & 24.1 & 3.8 & 11.9 & 47.4 & 43.1 & 11.4 & 16.9 & 30.6 & 7.2 & 27.5 \\
		Geiger and Wang~\cite{geiger2015joint} (NYU only) & 10.2 & 62.5 & 19.1 & 5.8 & 8.5 & 40.6 & 27.7 & 7.0 & 6.0 & 22.6 & 5.9 & 19.6 \\
		SSCNet~\cite{song2017semantic} & 15.1 & 94.6 & 24.7 & 10.8 & 17.3 & 53.2 & 45.9 & 15.9 & 13.9 & 31.1 & 12.6 & 30.5 \\
		VVNet~\cite{guo2018view} & 19.3 & \textbf{94.8} & 28.0 & 12.2 & \textbf{19.6} & 57.0 & 50.5 & 17.6 & 11.9 & 35.6 & 15.3 & 32.9 \\
		SaTNet~\cite{liu2018nips} & 17.3 & 92.1 & 28.0 & 16.6 & 19.3 & 57.5 & \textbf{53.8} & 17.7 & 18.5 & 38.4 & 18.9 & 34.4 \\
		\textbf{Proposed Method} & 36.2 & 93.8 & \textbf{29.2} & 18.9 & 17.7 & \textbf{61.6} & 52.9 & \textbf{23.3} & \textbf{19.5} & 45.4 & \textbf{20.0} & \textbf{37.1} \\
		-- \emph{without completion branch} & 35.8 & 94.1 & 28.9 & \textbf{19.2} & 16.8 & 61.4 & 53.5 & 23.0 & 14.0 & \textbf{45.6} & 18.9 & 36.5 \\
		-- \emph{without scene consistency} & \textbf{36.8} & 91.7 & 28.0 & 18.3 & 8.3 & 58.8 & 49.5 & 13.0 & 16.7 & 42.6 & 17.6 & 34.7 \\
	\bottomrule
	\end{tabular}
}
	\setlength{\abovecaptionskip}{\RemoveAboveCaptionTab}
\caption{Semantic scene completion results on the NYU test set with depth map for IoU (in \%). 
\label{tab:nyu_iou}}
\end{table*}

\begin{table}[t]
\centering
\begin{tabular}{
	l|
	>{\centering\arraybackslash}p{1.3cm}|
	>{\centering\arraybackslash}p{1.3cm}}
	\toprule
		\multicolumn{1}{c}{} &
		\multicolumn{1}{c}{SUNCG} &
		\multicolumn{1}{c}{NYU} \\
	\midrule
		Lin \etal~\cite{lin2013holistic} & -- & 36.4  \\
	3D-RecGAN~\cite{yang2018dense} & 72.1 & 51.3 \\
	Geiger and Wang~\cite{geiger2015joint} & -- & 44.4 \\
	SSCNet~\cite{song2017semantic} & 73.5 & 56.6 \\
	VVNet~\cite{guo2018view} & 84.0 & 61.1  \\
	SaTNet~\cite{liu2018nips} & 78.5 & 60.6 \\
	\textbf{Proposed Method} & \textbf{86.9} & \textbf{63.4} \\
	-- \emph{without completion branch} & 82.3 & 62.6   \\
	-- \emph{without scene consistency} & 82.0 & 61.1 \\
	\bottomrule
	\end{tabular}
	\setlength{\abovecaptionskip}{\RemoveAboveCaptionTab}
	\setlength{\belowcaptionskip}{\RemoveBelowCaption}
	\caption{Scene completion results on the SUNCG and the NYU test set in terms of IoU (in $\%$). \label{tab:completion_iou}}
\end{table}
\section{Experiments}

There are two tasks at hand -- (1)~3D semantic scene completion; and, (2)~3D object completion. Although they perform similar tasks in reconstructing from a single view, 
the former completes the structure of a scene with semantic labels while the latter requires a more detailed completion with the assumption of a single category.

\vspace{\RemoveAboveParagraph}
\paragraph{Metric.} 
For each of the $N$ classes, the accuracy of the predicted volumes is measured based on the Intersection over Union (IoU). 
Analogously to the evaluation carried out by other methods, the average IoU is taken from all the categories except for the empty space.

\vspace{\RemoveAboveParagraph}
\paragraph{Implementation details.} 
We learn our model with an Nvidia Titan Xp with a batch size of 8. 
We applied batch normalization after every convolutional and deconvolutional operations except for the convolutional operations in the last deconvolutional layers in 3 generators.
Leaky ReLU with a negative slope of 0.2 is applied on the output of each convolutional layer in the $\text{Res3D}(\cdot,\cdot)$ modules in \figref{fig:arch_micro}. In addition, ReLU is applied on the output of deconvolutional operations in the generators except for the last deconvolution operation in the Multi-Scale Upsampling. Finally, the sigmoid operation is applied to the last deconvolution layer of the generators for the geometric and semantic completion.
Notably, the factor $\lambda$ from \eqref{eq:per_category_error} is set to be 0.5 for the geometric completion in \eqref{eq:l_recon}.
For the semantic completion, it is initially set to 0.9 in \eqref{eq:l_pred}. However, when the network is capable of revealing objects from the depth image, more and more false positive predictions in the empty space appears. Due to this, we set $\lambda$ to 0.6 after five epochs.

\subsection{Semantic scene completion}
\label{sec:eval_scene}

The SUNCG~\cite{song2017semantic} and NYU~\cite{SilbermanECCV12} datasets are currently the most relevant benchmarks for semantic scene completion, and include a paired depth image and the corresponding semantically labeled volume.
While SUNCG comprises synthetically rendered depth data, NYU includes real scenes acquired with a Kinect depth sensor. 
This makes the evaluation of NYU more challenging, due to the presence of real nuisances, as well as due to a limited training set of less than 1000 samples.
We compare our method against
Wang \etal~\cite{wang3dv}, 
Lin \etal~\cite{lin2013holistic}, 
3D-RecGAN~\cite{yang2018dense},
Geiger and Wang~\cite{geiger2015joint},
SSCNet~\cite{song2017semantic}, 
VVNet~\cite{guo2018view}, and
SaTNet~\cite{liu2018nips}.
The resolution of our input volume is given in the scale of \dimthree{80}{48}{80} voxels. 
While \cite{geiger2015joint,guo2018view,lin2013holistic,liu2018nips,song2017semantic} produce \dimthree{60}{36}{60} semantic volumes for evaluation, 
\cite{wang3dv,yang2018dense} and us produce a slightly higher resolution of \dimthree{80}{48}{80}.

Following SUNCG~\cite{song2017semantic}, the semantic categories include 12 classes of varying shapes and sizes, \ie: \emph{empty space}, \emph{ceiling}, \emph{floor}, \emph{wall}, \emph{window}, \emph{chair}, \emph{bed}, \emph{sofa}, \emph{table}, \emph{tvs}, \emph{furniture} and \emph{other objects}. 
%
%
We follow two types of evaluation as introduced by \cite{song2017semantic}. 
One evaluates the semantic segmentation accuracy on the observed surface reconstruction, while the other considers the semantic segmentation of the predicted full volumetric reconstruction.

\vspace{\RemoveAboveParagraph}
\paragraph{SUNCG dataset.}

Based on an online interior design platform, the evaluation of SUNCG contains more than 130,000 paired depth images and voxel-wise semantic labels taken from 45,622 houses with realistic rooms and furniture layouts~\cite{song2017semantic}.
Focusing on the semantic segmentation on the observed surface, our approach performs at an IoU of 57.2\% which is 3.0\% higher than SSCNet~\cite{song2017semantic}. 
On the other hand, when we evaluate the IoU measure on the entire volume in \tabref{tab:suncg_iou}, our method reaches an average IoU of 63.4\% which is significantly better than 
Wang \etal~\cite{wang3dv}, 
3D-RecGAN~\cite{yang2018dense} and 
SSCNet~\cite{song2017semantic}
but slightly worse than VVNet~\cite{guo2018view} and SaTNet~\cite{liu2018nips}.

\vspace{\RemoveAboveParagraph}
\paragraph{NYU dataset (real).}

The NYU dataset~\cite{SilbermanECCV12} is composed of 1,449 indoor depth images captured with a Kinect depth sensor. Like SUNCG, each image is also annotated with 3D semantic labels. 
Due to its size, training our network on this dataset alone is insufficient. 
As a solution already used in~\cite{song2017semantic}, we take the network trained on the SUNCG then refine it by supplementing the training data from NYU with 1,500 randomly selected samples from SUNCG in each epoch of training.

Although we achieved slightly worse results than VVNet~\cite{guo2018view} and SaTNet~\cite{liu2018nips} on the synthetic dataset, we performed better than the state of the art on the real images,  reaching an IoU measure of 37.1\% as shown in \tabref{tab:nyu_iou}.
Consequently, we attain a 4.2\% improvement compared to VVNet~\cite{guo2018view} and 2.7\% to SaTNet~\cite{liu2018nips}.

Looking at the other approaches, we achieve even more significant improvements with at least 6.6\% increase in IoU.
For the evaluation on the semantic labels on the observed surface, we gained 4.4\% increase in IoU against SSCNet. 
Notably, our approach outperforms other works not only on the average IoU but also on individual object categories.
In addition, we also achieve similar improvements in the scene completion task in \tabref{tab:completion_iou} with approximately 2.8\% better in IoU compared to SaTNet~\cite{liu2018nips}.

Moreover, while the re-implementation SSCNet~\cite{song2017semantic} in our experiments does not fit into any of our contributions, we used it in order to qualitatively compare our results with them (see \figref{fig:teaser}).


\vspace{\RemoveAboveParagraph}
\paragraph{Ablation study for loss terms.}

In Tables~\ref{tab:suncg_iou} and~\ref{tab:nyu_iou}, we investigate the contribution of $\mathcal{L}_{\text{recon}}$ from the supervised learning and $\mathcal{L}_{\text{consistency}}$ from the unsupervised learning.
Our ablation study indicates that $\mathcal{L}_{\text{consistency}}$ prompts the highest boost in IoU with 5.2\% in \tabref{tab:suncg_iou}. 
When using the $\mathcal{L}_{\text{recon}}$ in the geometric completion, it improves by 1.1\% on the SUNCG dataset. 
A similar conclusion for the loss terms is presented in \tabref{tab:suncg_iou} for NYU.

\begin{table}[t]
\centering
\resizebox{\linewidth}{!}
{\begin{tabular}{l|cccc|c}
	\toprule	
		\multicolumn{1}{c}{} 
	& bench & chair & couch & table & \emph{Avg.} \\
	\midrule
	Varley \etal~\cite{varley2017shape} & 65.3 & 61.9 & 81.8 & 67.8 & 69.2 \\
	3D-EPN~\cite{dai2017shape} & 75.8 & 73.9 & 83.4 & 77.2 & 77.6 \\
	Han \etal~\cite{han2017high} & 54.4 & 46.9 & 48.3 & 56.0 & 51.4 \\
	3D-RecAE~\cite{yang2018dense} & 73.3 & 73.6 & 83.2 & 75.0 & 76.3 \\
	3D-RecGAN~\cite{yang2018dense} & 74.5 & 74.1 & 84.4 & 77.0 & 77.5 \\
	\textbf{Proposed Method} & \textbf{79.1} & \textbf{80.6} & \textbf{92.4} & \textbf{84.0} & \textbf{84.1} \\
	-- \emph{without scene consistency} & 76.3 & 76.4 & 87.5 & 81.2 & 80.4 \\
	\bottomrule
	\end{tabular}
}
	\setlength{\abovecaptionskip}{\RemoveAboveCaptionTab}
	\setlength{\belowcaptionskip}{\RemoveBelowCaption}
	\caption{Object completion results on the ShapeNet test set in terms of IoU (in $\%$). The resolution for Varley \etal~\cite{varley2017shape} and 3D-EPN~\cite{dai2017shape}: \dimthree{32}{32}{32}, for others: \dimthree{64}{64}{64}. \label{tab:shapenet_iou}}
\end{table}

\begin{table}[t]
\centering
\resizebox{\linewidth}{!}
{\begin{tabular}{l|cccc|c}
	\toprule
		\multicolumn{1}{c}{} 
	& bench & chair & couch & table & \emph{Avg.} \\
	\midrule
	Han \etal~\cite{han2017high} & 18.4 & 14.8 & 10.1 & 12.6 & 14.0 \\
	3D-RecAE~\cite{yang2018dense} & 23.1 & 17.8 & 10.7 & 14.8 & 16.6 \\
	3D-RecGAN~\cite{yang2018dense} & 23.0 & 17.4 & 10.9 & 14.6 & 16.5 \\
	\textbf{Proposed Method} & \textbf{32.7} & \textbf{24.1} & \textbf{15.9} & \textbf{22.5} & \textbf{23.8} \\
	-- \emph{without scene consistency} & 26.1 & 21.5 & 14.9 & 18.6 & 20.3 \\
	\bottomrule
	\end{tabular}
}
	\setlength{\abovecaptionskip}{\RemoveAboveCaptionTab}
	\setlength{\belowcaptionskip}{\RemoveBelowCaption}
\caption{Object completion results on the real-world  test set provided by 3D-RecGAN~\cite{yang2018dense} in terms of IoU (in $\%$). The resolution for all methods is \dimthree{64}{64}{64}. \label{tab:recgan_iou}}
\end{table}


\subsection{3D object completion}

Adapting the assessment data and strategy from 3D-RecGAN~\cite{yang2018dense}, we use ShapeNet~\cite{chang2015shapenet} to generate the training and test data for 3D object completion, wherein each reconstructed object surface $\inputsurface$ is paired with a corresponding ground truth voxelized shape with a size of \dimthree{64}{64}{64}. 
The dataset comprises four object classes: \emph{bench}, \emph{chair}, \emph{couch} and \emph{table}. \cite{yang2018dense} prepared an evaluation for both synthetic and real input data.
Notably, for both synthetic and real test data, we can express the same conclusions as the ablation studies in \secref{sec:eval_scene} (see Tables~\ref{tab:shapenet_iou} and~\ref{tab:recgan_iou}).

\vspace{\RemoveAboveParagraph}
\paragraph{Synthetic test data.}

We perform two evaluations in \tabref{tab:shapenet_iou}. The first is a single category test~\cite{yang2018dense} such that each category is trained and tested separately while the second considers the categories in order to label the voxels. We compare our results against \cite{dai2017shape,han2017high,varley2017shape,yang2018dense}.

In the single category test, we achieve the best results with 84.1\%. 
This result is 
6.5\% higher than 3D-EPN~\cite{dai2017shape}, 
6.6\% higher than 3D-RecGAN~\cite{yang2018dense}, 
7.8\% higher than 3D-RecAE~\cite{yang2018dense}, 
32.7\% higher than Han \etal~\cite{han2017high} and
14.9\% higher than Varley \etal~\cite{varley2017shape}. 
Moreover, this table also shows the we achieve the best results across all categories.

\vspace{\RemoveAboveParagraph}
\paragraph{Real test data.}

Using the single category test in \tabref{tab:recgan_iou}, we also evaluate the 3D object completion task on the real world test data provided by \cite{yang2018dense}.
In this evaluation, we generate the state-of-the art results with 23.8\% IoU measure, which is higher than 3D-RecAE~\cite{yang2018dense} by 7.2\%,
3D-RecGAN~\cite{yang2018dense} by 7.3\% and 
Han \etal~\cite{han2017high} by 9.8\%. 
%

\section{Conclusion}
We propose ForkNet, a novel architecture for volumetric semantic 3D completion that leverages a shared embedding encoding both geometric and semantic surface cues, as well as multiple generators designed to deal with limited paired data and imprecise semantic annotations. 
Experimental results numerically demonstrate the benefits of our approach for the two tasks of scene and object completion, as well as the effectiveness of the proposed contributions in terms of architecture, loss terms and use of discriminators. 
%
%
However, since we compress the input SDF volume into a lower resolution through the encoder then increase the resolution through the generator, small or thin structures such as the legs of the chair or TVs tend to disappear during compression. This is an aspect we plan to improve in the future work. 
In addition, for 3D scene understanding, the volumetric representations are typically memory and power-hungry, we also plan to extend our model for completion of efficient and sparse representations such as point clouds.  

\section*{Acknowledgment}


The authors would like to thank Shun-Cheng Wu for the
fruitful discussions and support in preparation of this work.

{\small
\bibliographystyle{ieee_fullname}
\bibliography{egpaper_final}
}

\end{document}